\documentclass{bmvc2k}
\usepackage{adjustbox}
\usepackage{multirow}
\usepackage{graphicx}
\usepackage{blindtext}
\usepackage{amsmath}
\usepackage{mathtools, nccmath}

\title{Explanation based Handwriting Verification}

\addauthor{Mihir Chauhan}{mihirhem@buffalo.edu}{1}
\addauthor{Mohammad Abuzar Shaikh}{mshaikh2@buffalo.edu}{1}
\addauthor{Sargur N. Srihari}{srihari@buffalo.edu}{2}

\addinstitution{
 State University of New York at Buffalo\\
 Buffalo, New York, USA
}

\runninghead{Chauhan, Shaikh \etal}{Explanation based Handwriting Verification}

\def\etal{\emph{et al}\bmvaOneDot}

\begin{document}

\maketitle

\begin{abstract}

Deep learning system have drawback that their output is not accompanied with explanation. In a domain such as forensic handwriting verification it is essential to provide explanation to jurors. The goal of handwriting verification is to find a measure of confidence whether the given handwritten samples are written by the same or different writer. We propose a method to generate explanations for the confidence provided by convolutional neural network (CNN) which maps the input image to 15 annotations (features) provided by experts. Our system comprises of: (1) Feature learning network (FLN), a differentiable system, (2) Inference module for providing explanations. Furthermore, inference module provides two types of explanations: (a) Based on cosine similarity between categorical probabilities of each feature, (b) Based on Log-Likelihood Ratio (LLR) using directed probabilistic graphical model. We perform experiments using a combination of feature learning network (FLN) and each inference module. We evaluate our system using \href{http://betelgeuse.cse.buffalo.edu/and_dataset}{XAI-AND} dataset, containing 13700 handwritten samples and 15 corresponding expert examined features for each sample. The dataset is released for public use and the methods can be extended to provide explanations on other verification tasks like face verification and bio-medical comparison. This dataset can serve as the basis and benchmark for future research in explanation based handwriting verification. The code is available on \href{https://github.com/mshaikh2/HDL_Forensics}{github}.
\end{abstract}
\section{Introduction}
\label{sec:intro}
Handwritten evidences provided by expert forensic document examiners (FDE) has long been admissible in the court of law. FDE subjectively specify the characteristics of the handwritten samples like word formations, pen pressure and pen lifts which uniquely identifies an individual writer. The premise for finding unique characteristics is based on the hypothesis that every individual has a unique way of writing \cite{Individuality:1}. The examination of the handwritten samples involves comparison of the questioned handwritten sample submitted for examination with the known handwritten sample under investigation. Therefore, the task of handwriting verification is to find a measure of confidence whether the questioned and known handwritten samples are written by the same writer or different writer. An example of handwriting verification evidence as presented by FDE is shown in Figure 1. \cite{gunning2017explainable}
\begin{figure}[htb!]
\centering
\includegraphics[scale=0.19]{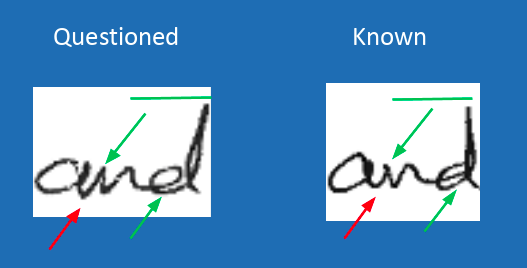}
\caption{Explanation with evidence provided to the court of law by a QD examiner. Red arrow indicates a dissimilarity in staff of 'a'. The two green arrows indicates similarity between staff of 'n' and exit stroke of 'd' respectively. A green bar over d indicates similarity between staff of 'd'. The explanations provides a confidence that the two handwritten samples were written by same writer.}
\end{figure}

Forensic handwritten evidences has received skepticism on the reliability of the reasoning methodologies and the subjective nature of the judgments provided by FDE. This is due to the non-exact nature of the conclusions drawn from the explanations provided by different FDE's. 

Researchers have implemented handwriting verification systems using conventional machine learning techniques \cite{intro3:14} \cite{Individuality:1} \cite{intro1:12} as well as deep learning techniques \cite{HybridFeatureLearning} \cite{chu2018writer}. FDE's are still unconvinced with the handwriting verification systems because: (1) The output of such systems is difficult to interpret because the system does not provide explanations for the decision. (2) The system is opaque and the inner-working of the system is unclear. Hence, our goal is to provide an explanation based handwriting verification system which provides a decision report for the FDE to interpret the output. Vantage writer approach \cite{brink2007towards} was proposed to generate comprehensible reports for providing explanations for the task of writer verification and identification. The idea used in \cite{brink2007towards} is that an individual handwriting can be seen as a mixture of handwritings of typical writers, the vantage writers. The vantage profile describes the features for a handwritten sample. The profile is created using similarity between features of a handwritten sample and all vantage handwritten samples. We have proposed a different approach wherein we generate human explainable features instead of using vantage profile. The proposed system consists of two modules:(1) Feature Learning Network (FLN) which learns to map the input images to expert human observed features (2) The decision inference interface which uses cosine similarity and probabilistic graphical network to provide rational behind the decision made by the system. In the next section, we describe XAI-AND, a dataset specifically created for explainable handwriting verification. \vspace{-0.8em}

\begin{table}[ht]
\begin{center}
\begin{tabular}{|l|l|}
\hline
\textbf{Element} & \textbf{Explainable Features} \\ \hline
Form & Entry/Exit Stroke \\ \hline
Pressure & Pen Pressure \\ \hline
Speed & Constancy \\ \hline
Dimension & Word Dimension, Word Size \\ \hline
Continuity & Is Cursive \\ \hline
Direction & Tilt, Slantness \\ \hline
Order & Formation \\ \hline
\end{tabular}
\end{center}
\caption{Seven Elements of Handwriting}
\vspace{-1.5em}
\end{table}

\begin{table}[]
\centering
\begin{center}
\begin{tabular}{|l|l|l|l|l|}
\hline
\textbf{Pen Pressure} & \textbf{Tilt} & \textbf{Entry Stroke of "a"} & \textbf{Is Lowercase} & \textbf{Is Continuous}\\ \hline \hline
\multicolumn{1}{|c|}{\begin{tabular}[c]{@{}c@{}}\includegraphics[scale=0.3]{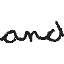}\\ Strong \\ (40.6\%)\end{tabular}} & 
\multicolumn{1}{|c|}{\begin{tabular}[c]{@{}c@{}}\includegraphics[scale=0.3]{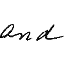}\\ Normal\\ (81.24\%)\end{tabular}} &
\multicolumn{1}{|c|}{\begin{tabular}[c]{@{}c@{}}\includegraphics[scale=0.3]{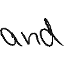}\\ No Stroke \\ (94.32\%) \end{tabular}} &
\multicolumn{1}{|c|}{\begin{tabular}[c]{@{}c@{}}\includegraphics[scale=0.3]{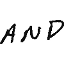}\\ No \\ (1.5\%)\end{tabular}} & 
\multicolumn{1}{|c|}{\begin{tabular}[c]{@{}c@{}}\includegraphics[scale=0.3]{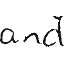}\\ No \\ (33.38\%)\end{tabular}} \\ \hline

\multicolumn{1}{|c|}{\begin{tabular}[c]{@{}c@{}}\includegraphics[scale=0.3]{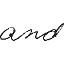}\\ Medium \\ (59.4\%)\end{tabular}} & 
\multicolumn{1}{|c|}{\begin{tabular}[c]{@{}c@{}}\includegraphics[scale=0.3]{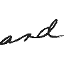}\\ Tilted \\ (18.76\%)\end{tabular}} &
\multicolumn{1}{|c|}{\begin{tabular}[c]{@{}c@{}}\includegraphics[scale=0.3]{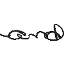}\\ Downstroke \\ (5.68\%)\end{tabular}} &
\multicolumn{1}{|c|}{\begin{tabular}[c]{@{}c@{}}\includegraphics[scale=0.3]{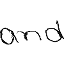}\\ Yes\\ (98.5\%)\end{tabular}} & 
\multicolumn{1}{|c|}{\begin{tabular}[c]{@{}c@{}}\includegraphics[scale=0.3]{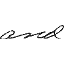}\\ Yes\\ (66.62\%)\end{tabular}} \\ \hline
\end{tabular}
\end{center}
\caption{Examples and Class Distribution}
\end{table}

\section{Dataset}
XAI-AND dataset is a publicly available dataset for handwriting verification, comprising of 15,518 ``AND'' image fragments extracted from CEDAR Letter Dataset \cite{Individuality:1} written by 1567 writers. Each ``AND'' image fragment is labeled by a questioned document (QD) examiner with 15 explainable discrete features. QD examiners have specified these handwriting features based on years of training using seven fundamental elements of handwriting \cite{sevenrules} as shown in Table 1. We have created a web based truthing tool for QD examiners to enter the values for the 15 features for ``AND'' images fragments. The data entry work using the truthing tool was shared primarily between 89 external examiners. The data entered by the external examiners was verified by 2 QD examiners. The resultant dataset serves as a good resource for explanation based handwriting verification. Table 2, 3 and 4 shows class-wise examples and distribution for each explainable feature. \\ \\
\textbf{Data Partitioning} We follow three approaches to partitioning the writers in training ($D_{train}$), validation ($D_{val}$) and testing($D_{test}$) set \cite{HybridFeatureLearning}:
\newline
\textit{Unseen Writer Partitioning}: No data partitions share any sample from same writer\useshortskip
\begin{equation}
D_{train} \bigcap D_{val} \bigcap D_{test} = \emptyset \label{eq}
\vspace{-1mm}
\end{equation}
\textit{Shuffled Writer Partitioning}: Data partitions randomly share different samples from writers\useshortskip
\begin{equation}
D_{train} \bigcap D_{val} \bigcap D_{test} = X \quad \text{where} \quad X \neq \emptyset \label{eq}
\vspace{-1mm}
\end{equation}
\textit{Seen Writer Partitioning}: Each data partition contains different samples of the same writer\useshortskip
\begin{equation}
D_{train} =  \bigcup_{i = 1}^{N} 0.6 * S_{i}, \enspace
D_{val} =  \bigcup_{j = 1}^{N} 0.2 * S_{j}, \enspace
D_{test} =  \bigcup_{k = 1}^{N} 0.2 * S_{k}, \enspace
D_{train} \bigcup D_{val} \bigcup D_{test} = S \label{eq} 
\end{equation}
where $S$ denotes a set of all the writers and $i, j, k$ denote different samples from each writer

\begin{table}[]
\centering
\begin{center}
\begin{tabular}{|l|l|l|l|l|l|}
\hline
\textbf{Slantness} & \textbf{Size} & \textbf{Staff of 'a'} & \textbf{Dimension} & \textbf{Exit Stroke 'd'}\\ \hline \hline 
\multicolumn{1}{|c|}{\begin{tabular}[c]{@{}c@{}}\includegraphics[scale=0.3]{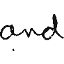}\\ Normal \\ (52.41\%) \end{tabular}} &
\multicolumn{1}{|c|}{\begin{tabular}[c]{@{}c@{}}\includegraphics[scale=0.3]{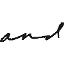}\\ Small \\ (23.01\%)\end{tabular}} &
\multicolumn{1}{|c|}{\begin{tabular}[c]{@{}c@{}}\includegraphics[scale=0.3]{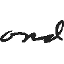} \\ No Staff \\ (18.04\%)\end{tabular}} &
\multicolumn{1}{|c|}{\begin{tabular}[c]{@{}c@{}}\includegraphics[scale=0.3]{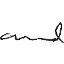}\\ Low \\ (29.75\%) \end{tabular}} & 
\multicolumn{1}{|c|}{\begin{tabular}[c]{@{}c@{}}\includegraphics[scale=0.3]{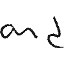}\\ No Stroke \\ (24.86\%) \end{tabular}} \\ \hline

\multicolumn{1}{|c|}{\begin{tabular}[c]{@{}c@{}}\includegraphics[scale=0.3]{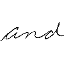}\\ Slight Right \\ (29.38\%) \end{tabular}} &
\multicolumn{1}{|c|}{\begin{tabular}[c]{@{}c@{}}\includegraphics[scale=0.3]{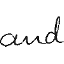}\\ Medium \\ (52.41\%) \end{tabular}} &
\multicolumn{1}{|c|}{\begin{tabular}[c]{@{}c@{}}\includegraphics[scale=0.3]{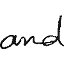}\\ Retraced \\ (58.45\%) \end{tabular}} &
\multicolumn{1}{|c|}{\begin{tabular}[c]{@{}c@{}}\includegraphics[scale=0.3]{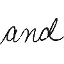}\\ Medium \\ (52.18\%) \end{tabular}} & 
\multicolumn{1}{|c|}{\begin{tabular}[c]{@{}c@{}}\includegraphics[scale=0.3]{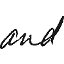}\\ Down Stroke \\ (44.02\%)\end{tabular}} \\ \hline

\multicolumn{1}{|c|}{\begin{tabular}[c]{@{}c@{}}\includegraphics[scale=0.3]{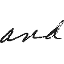}\\ Very Right \\ (11.05\%) \end{tabular}} &
\multicolumn{1}{|c|}{\begin{tabular}[c]{@{}c@{}}\includegraphics[scale=0.3]{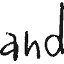}\\ Large \\ (24.58\%)\end{tabular}} &
\multicolumn{1}{|c|}{\begin{tabular}[c]{@{}c@{}}\includegraphics[scale=0.3]{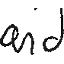}\\ Loopy \\ (7\%)\end{tabular}} &
\multicolumn{1}{|c|}{\begin{tabular}[c]{@{}c@{}}\includegraphics[scale=0.3]{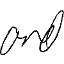}\\ High \\ (18.07\%)\end{tabular}} & 
\multicolumn{1}{|c|}{\begin{tabular}[c]{@{}c@{}}\includegraphics[scale=0.3]{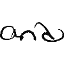}\\ Curved Up \\ (12.6\%)\end{tabular}}
 \\ \hline
\multicolumn{1}{|c|}{\begin{tabular}[c]{@{}c@{}}\includegraphics[scale=0.3]{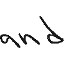}\\ Left \\ (7.58\%) \end{tabular}} &
 &
\multicolumn{1}{|c|}{\begin{tabular}[c]{@{}c@{}}\includegraphics[scale=0.3]{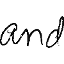}\\ Tented \\ (16.51\%)\end{tabular}} &
 & 
\multicolumn{1}{|c|}{\begin{tabular}[c]{@{}c@{}}\includegraphics[scale=0.3]{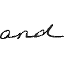}\\ Straight \\ (18.53\%)\end{tabular}}
 \\ \hline
\end{tabular}
\end{center}
\caption{Examples and Class Distribution (contd.)}
\end{table}

\section{Methods}
We propose two broad approaches through our experiments. Both the approaches have two parts to it: (i) A deep learning \cite{lecun1995convolutional} model for feature extraction, (ii) Inference model for providing an explanation interface \cite{gunning2017explainable}. In the first approach we find cosine similarity \cite{mihalcea2006corpus} between the soft assigned class values of corresponding 15 features of two given images, then display the degree of similarity of each feature and overall matching score. In the second, we infer the degree of similarity between the hard assigned predicted class values of the two images through bayesian inference \cite{lindley}.

\begin{table}[t]
\centering
\begin{center}
\begin{tabular}{|l|l|l|l|l|}
\hline
\textbf{Constancy} & \textbf{Letter Spacing} & \textbf{Word Formation}  & \textbf{Staff of 'd'} & \textbf{Formation of 'n'}\\ \hline \hline
\multicolumn{1}{|c|}{\begin{tabular}[c]{@{}c@{}}\includegraphics[scale=0.3]{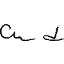}\\ Irregular\\ (39.65\%)\end{tabular}} & 
\multicolumn{1}{|c|}{\begin{tabular}[c]{@{}c@{}}\includegraphics[scale=0.3]{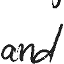}\\ Less\\ (22.49\%)\end{tabular}} &
\multicolumn{1}{|c|}{\begin{tabular}[c]{@{}c@{}}\includegraphics[scale=0.3]{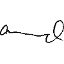}\\ Not Well Formed\\ (56.91\%)\end{tabular}} &
\multicolumn{1}{|c|}{\begin{tabular}[c]{@{}c@{}}\includegraphics[scale=0.3]{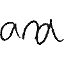}\\ No Staff\\ (9.86\%)\end{tabular}} &
\multicolumn{1}{|c|}{\begin{tabular}[c]{@{}c@{}}\includegraphics[scale=0.3]{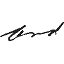}\\ No Formation\\ (22.97\%)\end{tabular}}\\ \hline

\multicolumn{1}{|c|}{\begin{tabular}[c]{@{}c@{}}\includegraphics[scale=0.3]{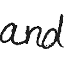}\\ Regular \\ (60.35\%)\end{tabular}} & 
\multicolumn{1}{|c|}{\begin{tabular}[c]{@{}c@{}}\includegraphics[scale=0.3]{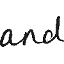}\\ Medium \\ (43.09\%)\end{tabular}} &
\multicolumn{1}{|c|}{\begin{tabular}[c]{@{}c@{}}\includegraphics[scale=0.3]{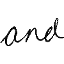}\\ Well Formed \\ (43.09\%)\end{tabular}} &
\multicolumn{1}{|c|}{\begin{tabular}[c]{@{}c@{}}\includegraphics[scale=0.3]{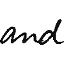}\\ Retraced \\ (49.63\%)\end{tabular}} &
\multicolumn{1}{|c|}{\begin{tabular}[c]{@{}c@{}}\includegraphics[scale=0.3]{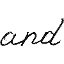}\\ Normal \\ (77.03\%)\end{tabular}} \\ \hline

 & 
\multicolumn{1}{|c|}{\begin{tabular}[c]{@{}c@{}}\includegraphics[scale=0.3]{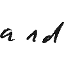}\\ High\\ (25.78\%)\end{tabular}} &
 &
\multicolumn{1}{|c|}{\begin{tabular}[c]{@{}c@{}}\includegraphics[scale=0.3]{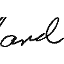}\\ Loopy\\ (40.51\%)\end{tabular}} &
 \\ \hline
\end{tabular}
\end{center}
\caption{Examples and Class Distribution (contd.)}
\end{table}
\subsection{Deep Feature Learning}
Various deep learning models were tested for mapping the human annotations to the handwritten image samples. For obtaining a baseline, we implemented a simple {\bf{deep CNN network}} to learn the mapping between input image and the 15 categories. The parameters of the deep CNN network are displayed in Figure 2 (a). This network comprises of an input of shape 64x64x1, layers from Encoding 1 to Encoding 5, followed by a FLN network to generate a multi-task output as shown in Figure 2 (b). Loss function for deep CNN network is shown in equation 4. However, since our input images were noisy, we reconstructed the input using an {\bf{auto-encoder}} (AE) in order generate clean output and improve effectiveness. The AE network resembles Figure 2, but, without any skip connections. Furthermore, to reduce the training time, avoid vanishing gradient and to recover a cleaner image, we added {\bf{skip connections}}  \cite{mao2016image} to the AE network. To make the reconstruction and learning non-trivial we translate the input data in the vertical direction and induce 50\% random noise in it. The task of reconstruction is to regenerate a clean and non-translated image.
    
Based on results of our experiments shown in Table 6, we employ {\bf{Skip Auto-Encoder}} (SAE) \cite{mao2016image} as it gives decent accuracy to demonstrate the main purpose of the paper. The input image is encoded through layers Encoding 1 up to Encoding 5 and then decoded through layers Decoding 4 down to 0. This is shown in Figure 1 (a). Each encoding layer consists of convolution \cite{lecun1995convolutional} followed by a max-pool operation. Furthermore, each convolution layer has a kernel size (3x3) and each max-pooling layer has a pool size (2x2). The number of filters keep increasing in multiples of 2 stating from 16 in Encoding 1. On decoding end, each layer consists of up-sampling followed by a convolution layer. Moreover, each up-sampling layer doubles the height and width of the incoming tensor and each conv. layer has kernel size (3x3). The number of filters keep decreasing by multiples of 2 starting from 256 in Decoding 4. We then concatenate the encoded feature maps with the decoding layers that have the same height and width to increase the robustness of training and reduce the effect of vanishing gradient problem \cite{mao2016image}. Thus, doubling the number of feature maps in the decoding layer. We believe the reconstruction of the noisy-translated samples as de-noised and centered samples provides a regularization effect for updating the parameters of FLN.
\begin{figure}[htbp]
\centering
\fbox{\includegraphics[scale=0.215]{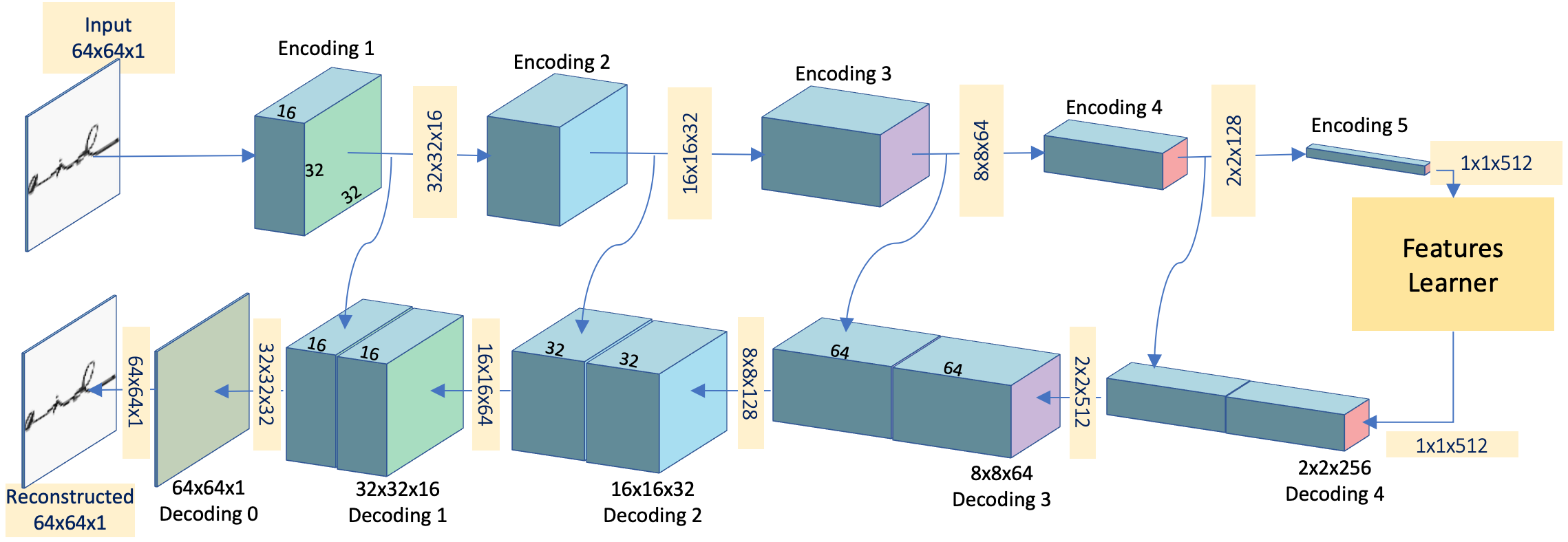}}
\linebreak
\label{ae}{(a)}
\linebreak
\fbox{\includegraphics[scale=0.215]{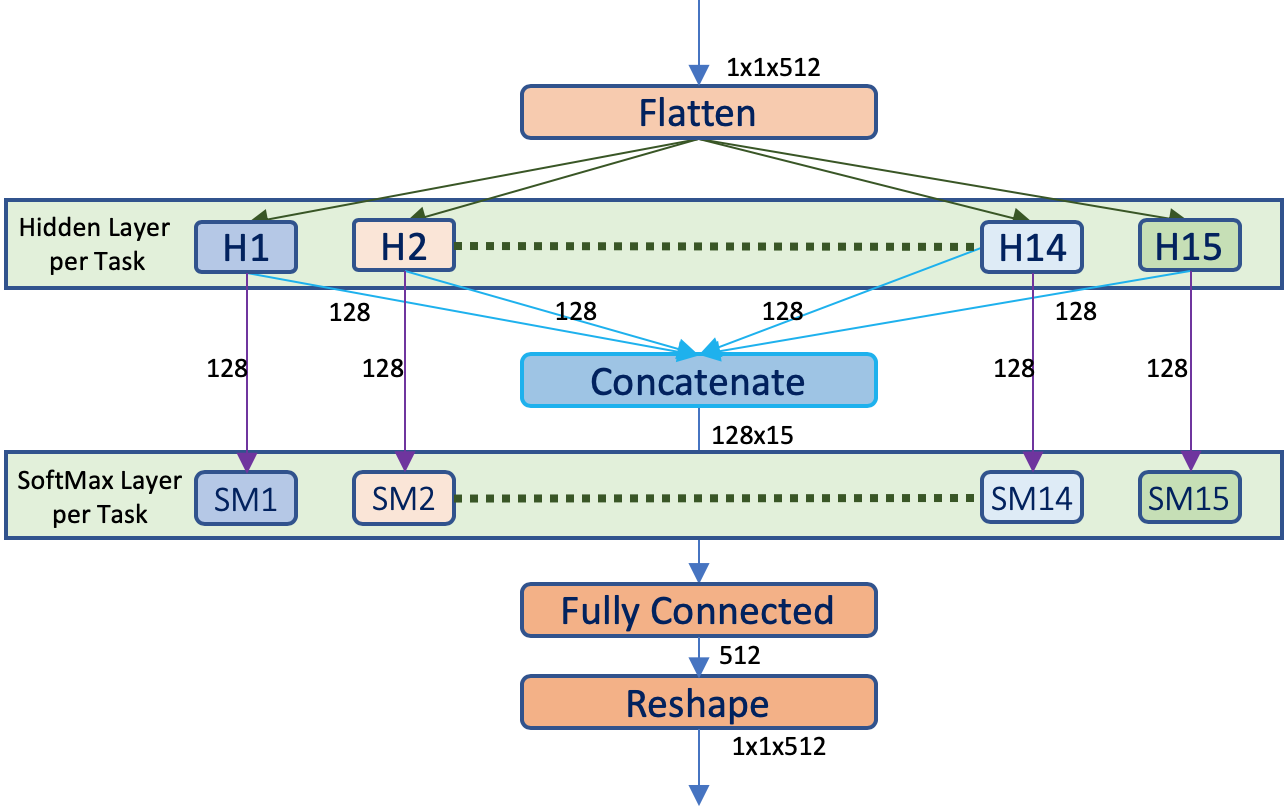}}
\linebreak\label{fln}{(b)}
\caption{(a) shows the Auto Encoder Architecture. (b) shows the expanded form of Features Learner we call it Feature Learner Network (FLN). In (b) H1 to H15 are hidden layers consisting 128 neurons for respective 15 tasks. SM1 to SM15 are soft-max activations for respective 15 tasks. Best viewed when zoomed}
\end{figure}

{\bf Feature Learner Network (FLN)}: We introduce a learning network such that Encoding 5 is processed in FLN and its output is then supplied to Decoding 4 for reconstruction. FLN consists of learning units for each of the 15 categories present in the dataset. Each learning unit comprises of a hidden layer (H) and a SoftMax layer (SM) such that there is one H and SM for each task. Each H has 128 neurons and each SM has neurons corresponding to the number of classes in respective task. The sum of Categorical-CrossEntropy (CCE) loss for each SM is backpropogated during the training. Hence the loss of FLN is denoted by:\useshortskip
\begin{equation}
  \label{eq:multipleentropy}
    L_{FLN} := \sum_{j} - \sum_{i} ({y_i' \log(y_i) + (1-y_i') \log (1-y_i)})
\vspace{-2mm}
\end{equation}
where $i$ is the class is each task category denoted in Table 2, 3 and 4. and $j$ ranges from 1 to 15 for each of the categories. Next, all the H's are concatenated and input to a fully connected (FC) layer consisting of 512 neurons. The output of FC is then reshaped to 1x1x512 and is then input to Decoding 4 layer.
Hence, instead of passing the Encoding 5 directly to Decoding 4, which normally is the case, we do the following: (i) fan out Encoding 5 into 15 clone  branches, (ii) learn mapping of each category with the image, as Encoding 5 is a latent representation of the image, (iii) combine the learned representations of each category to further reconstruct the input image.

{\bf{Total Loss}}: Total Loss of the deep learning network is the sum of Reconstruction Loss ($L_{Recon}$) and FLN Loss ($L_{FLN}$). Where $L_{Recon}$ is calculated by measuring the models ability to reconstruct the image close to original. To generate the image the Decoding 5 neurons are activated by Sigmoid function since the input was normalized to be between 0 and 1.
\begin{equation}
  \label{eq:binarycrossentropy}
    L_{Recon} := - \sum_{i=1}^m y_i ln(p_i) + (1-y_i) log (1-p_i)
\end{equation}
\begin{equation}
  \label{eq:totalloss}
    L_{Total} := L_{Recon} + L_{FLN}
\vspace{-2mm}
\end{equation}


\subsection{Inference Methods for Explainability}
Analysis of handwriting features helps a Forensic Document Examiner (FDE) to find the probability ($p$) that the known ($k_{f}$) and questioned ($q_{f}$) handwritten samples were written by the same writer. Each handwritten sample, $q_{f}$ and $k_{f}$ is associated with 15 discrete features $f = \{f_{1}, f_{2}, ... f_{15}\}$. We use $D_{train}$ for training the FLN; $D_{val}$ to tune model hyper-parameters and threshold value; $D_{test}$ for testing the model with tuned parameters. Furthermore, two approaches are tested for analysis of handwritten samples:

\textbf{Distance as a measure (DAAM):} We use Cosine Similarity ($C_{sim}$) to measure the degree of similarity. We measure the similarity of the categorical probabilities learned by the FLN soft-max layer. Cosine similarity between the corresponding categories of two input images implies the measure of similarity: \useshortskip
\begin{equation}
  \label{eq:cosinesimilarity}
C_{sim} (q_{f_{j}},k_{f_{j}}) = \frac{\sum_{i=1}^{n}{q_{f_{j_{i}}} k_{f_{j_{i}}}}}
           {\sqrt{\sum_{i=1}^{n}{q_{f_{j_{i}}}^{2}}}
            \sqrt{\sum_{i=1}^{n}{k_{f_{j_{i}}}^{2}}}}
\end{equation} 
where $j$ $\in$  $\{1, 2, ... 15\}$ and $n$ is the number of classes in $f_{j}$. We also compute the Overall Cosine Similarity $C_{OCS}$ by taking the mean of the sum of $C_{sim}$ across all $f$. \useshortskip

\begin{equation}
  \label{eq:cosinesimilarity}
C_{OCS} (q_{f},k_{f}) = \sum_{j=1}^{15}{C_{sim} (q_{f_{j}},k_{f_{j}})}
\end{equation} 
Once the model learns to map the input images to human features using $D_{train}$, we then finalize a threshold value $T$ on the validation set. To guesstimate the value of $T$ we run multiple iteration of same experiment with $T$ ranging from 0.1 to 0.9 at an increment of 0.1. During every iteration, $C_{OCS}$ is calculated for each pair of data point in $D_{val}$. Simultaneously, we calculate the True Positives (TP), False Positives (FP), True Negatives (TN), False Negatives (FN), for all the pairs in $D_{val}$ based on the current increment of $T$. Post every iteration Precision and Recall is calculated. Finally, the value of $T$ where precision is very close to recall is chosen for testing the performance on $D_{test}$. Furthermore, during testing if the $C_{OCS}$ score is below $T$; the two samples are considered as negative pairs else positive pairs.  

\textbf{Likelihood as a measure (LAAM):}
Likelihood ratio (LR) is the ratio of the joint probability $P(q_{f},k_{f}|l_{0})$ of $q_{f}$ and $k_{f}$ given the handwritten samples were written by the same writer $l_{0}$ to the joint probability $P(q_{f},k_{f}|l_{1})$ of $q_{f}$ and $k_{f}$ given the samples were written by different writers $l_{1}$. Finding $P(q_{f},k_{f}|l)$ requires calculating the joint probability of existence for all possible combinations pairs of input features. This is computationally expensive and infeasible. Hence, we simplify the calculation by calculating distance between $q_{f}$ and $k_{f}$. This approach has been proposed for shoe-print verification, fingerprint verification by Yi Tang \etal in \cite{lindley}. \useshortskip
\begin{equation}
LR = \frac{P(q_{f},k_{f}|l_{0})}{P(q_{f},k_{f}|l_{1})} \approx \frac{P(d(q_{f},k_{f})|l_{0})}{P(d(q_{f},k_{f})|l_{1})}
\end{equation}
The distance $d(q_{f},k_{f})$ can be regarded as a measure of similarity between $q_{f}$ and $k_{f}$. 
\begin{figure}[htbp]
\centering
\fbox{\includegraphics[scale=0.25]{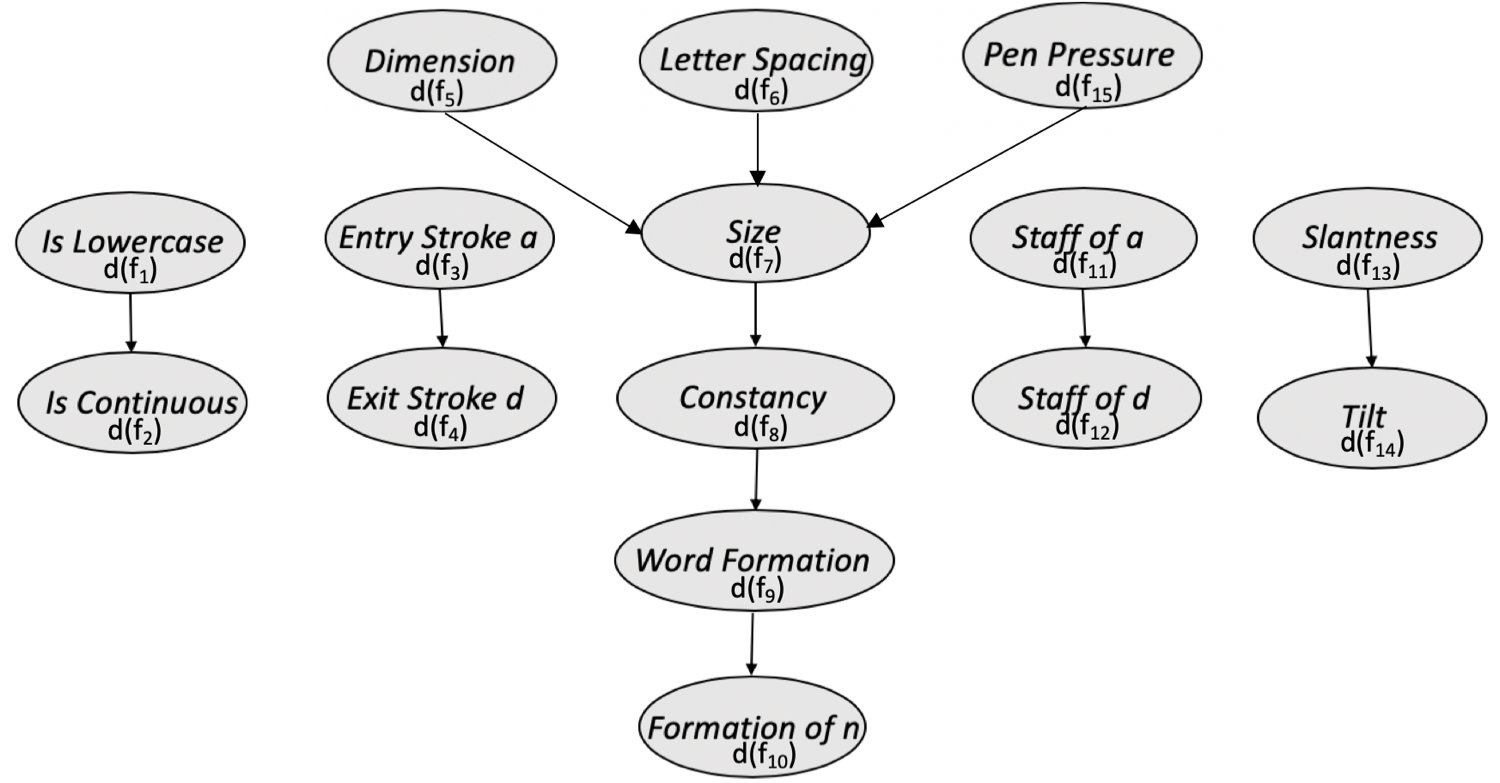}}
\caption{Bayesian network structure for difference distribution $P(d(q,k)|l)$}
\vspace{-2mm}
\end{figure}

Because each feature $f_{j} = [x_{1}, ... ,x_{n_{f_{j}}}]$ can take one of $n_{f_{j}}$ values, the features are multinomial in nature. In our dataset, $n_{f_{1}} = 2, n_{f_{2}} = 2, n_{f_{3}} = 2,n_{f_{4}} = 4,n_{f_{5}} = 3,n_{f_{6}} = 3,n_{f_{7}} = 3,n_{f_{8}} = 2,n_{f_{9}} = 2,n_{f_{10}} = 2,n_{f_{11}} = 4,n_{f_{12}} = 3,n_{f_{13}} = 4,n_{f_{14}} = 2, n_{f_{15}} = 2$. We have calculated the distance between multinomial discrete features using categorical distance values. Let's describe categorical distance by an example with staff of 'a' $f_{11}$ taking values \{0:No Staff, 1:Retraced, 2:Loopy, 3:Tented\}. Hence, $q_{f_{11}}$ and $k_{f_{11}}$ can take values \{0,1,2,3\}. As a result, $d(q_{f_{11}},k_{f_{11}})$ takes 10 categorical distance values: \{1:`00', 2:`01', 3:`02', 4:`03', 5:`11',6:`12', 7:`13', 8:`22', 9:`23', 10:`33'\}. We calculate the joint probability $P(d(q_{f},k_{f})|l_{0})$ using Bayesian Network ($BN1$) inference as in  eqn. 10 by considering dependencies between the 15 features as shown in Figure 3. Input to $BN1$ are the categorical distance values between $k_{f}$ and $q_{f}$ when $l = 0$. Similarly, we calculate the joint probability $P(d(q_{f},k_{f})|l_{1})$ using Bayesian Network $BN2$ inference using eqn. 10 while considering same dependencies between 15 features as shown in Figure 3. Input to $BN2$ are the categorical distance values between $k_{f}$ and $q_{f}$ when $l = 1$.

Both, BN1 and BN2 have 15 vertices with the same structure as shown in Figure 2. Furthermore, the structure is learned using correlation values, k2 scores, BDeu scores, BIC scores and domain knowledge. Each vertex contains a categorical distance value. Even though the structure of the Bayesian Network is the same for BN1 and BN2, we find that the conditional probability distributions (CPDs) generated using Maximum Likelihood Estimation (MLE) would be different. Finally, we infer the values of $P(d(q_{f},k_{f})|l_{0})$ and $P(d(q_{f},k_{f})|l_{1})$ using the CPDs from $BN1$ and $BN2$ respectively to calculate LR using equation [9][10]:
\begin{equation*}\label{eq:pareto}
\begin{aligned}
P(d(q_{f},k_{f})|l) &=   
P(d(f_{1}))*P(d(f_{2})|d(f_{1}))*P(d(f_{3}))*P(d(f_{4})|d(f_{3}))* \\
&\quad P(d(f_{11}))*P(d(f_{12})|d(f_{11}))*P(d(f_{13}))*P(d(f_{14})|d(f_{13}))* \\
&\quad P(d(f_{8})|d(f_{7}))*P(d(f_{9})|d(f_{8}))*P(d(f_{10})|d(f_{9}))*P(d(f_{5}))* \\
&\quad P(d(f_{6}))*P(d(f_{15}))*P(d(f_{7})|P(d(f_{5})),P(d(f_{6})),P(d(f_{15}))) \\
\quad \text{where,  } d(f_{j}) &= d(q_{f_{j}},k_{f_{j}})  \quad\quad\quad\quad\quad
\quad\quad\quad\quad\quad\quad\quad\quad\quad\quad\quad\quad\quad\quad\quad\quad\quad\quad\quad  (10)
\end{aligned}
\end{equation*}

\begin{figure}
\begin{tabular}{cc}
\bmvaHangBox{\includegraphics[scale=0.4]{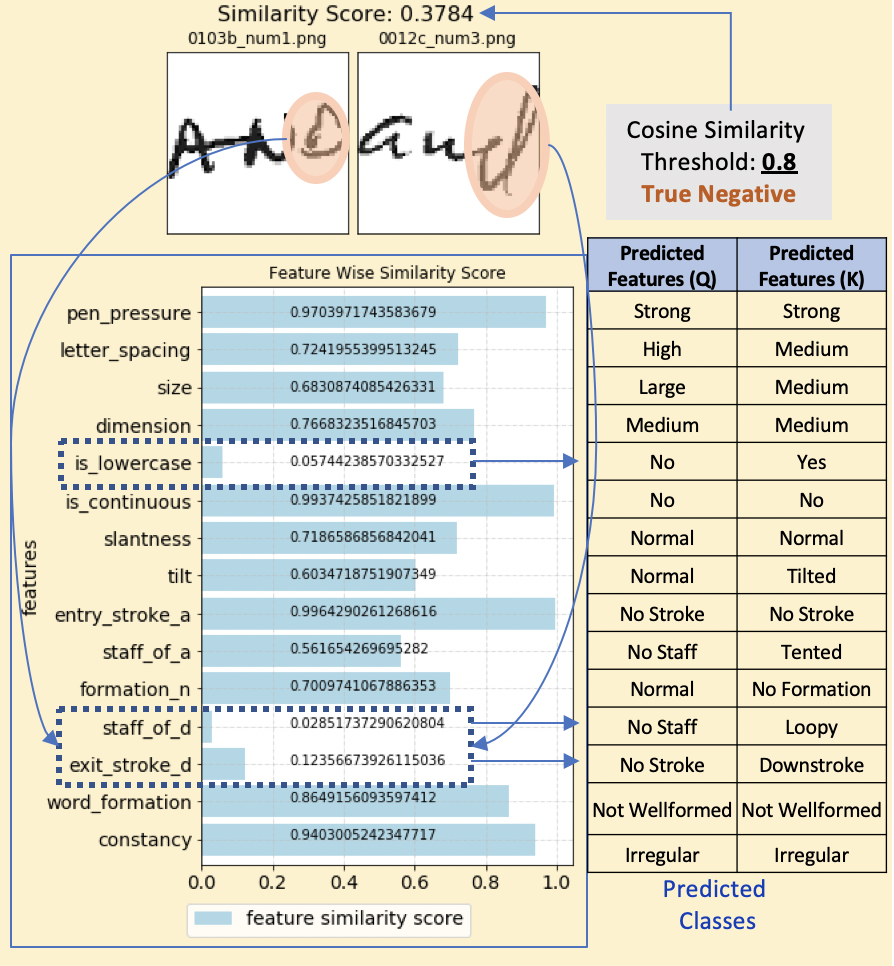}}&
\bmvaHangBox{\includegraphics[scale=0.4]{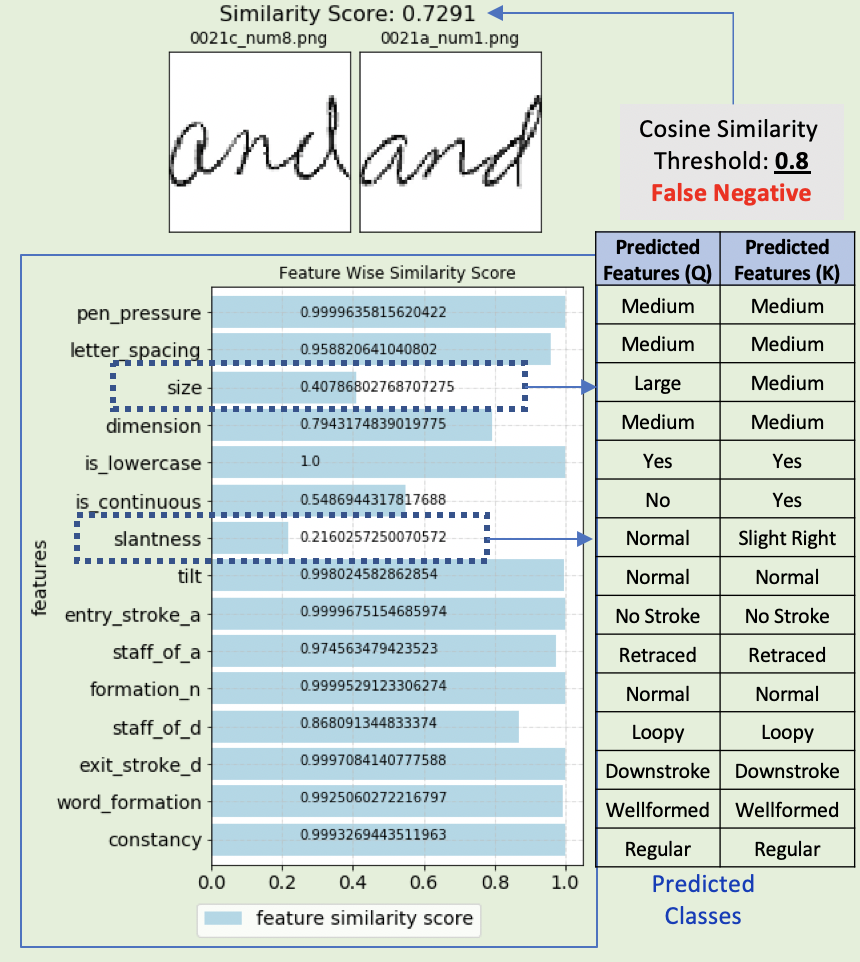}}
\\
(a) & (b)
\end{tabular}
\caption{Graph of Similarity Score of Questioned (q) \& Known (k) Image Features for (a) Different Writer Samples (b) Same Writer Samples
}
\end{figure}
\section{Experiments and Results}
Our experimental setup includes a deep learning system followed by an inference model. Training and testing of the deep learning system was done using three 11GB NVIDIA GTX 1080 Ti GPUs and TensorFlow backend. Input to the deep learning system are handwritten ``AND'' image snippets and the targets are the 15 explainable features. We normalize the input image snippets such that each pixel value in the image is between [0,1]. We develop data-generator to output 128 samples in one batch. The input image is translated in vertical direction randomly by $\pm$12px. This makes the skip auto-encoder translation invariant. We train the deep learning model for 100,000 epochs with SGD optimizer. The evaluation of the deep learning model is done using accuracy per explainable feature as shown in Table 5.

\begin{table}[ht]
\begin{center}
\begin{tabular}{|l|l|l|l|l|l|l|l|l|l|l|l|l|l|l|l|}
\hline
\textbf{\rotatebox{90}{Feature Name}} & \textbf{\rotatebox{90}{pen\_pressure}} & \textbf{\rotatebox{90}{letter\_spacing}} & \textbf{\rotatebox{90}{size}} & \textbf{\rotatebox{90}{dimension}} & \textbf{\rotatebox{90}{is\_lowercase}} & \textbf{\rotatebox{90}{is\_continuous}} & \textbf{\rotatebox{90}{slantness}} & \textbf{\rotatebox{90}{tilt}} & \textbf{\rotatebox{90}{entry\_stroke\_a}} & \textbf{\rotatebox{90}{staff\_of\_a}} & \textbf{\rotatebox{90}{formation\_n}} & \textbf{\rotatebox{90}{staff\_of\_d}} & \textbf{\rotatebox{90}{exit\_stroke\_d}} & \textbf{\rotatebox{90}{word\_formation}} & \textbf{\rotatebox{90}{constancy}} \\ 
\hline
\rotatebox{90}{Val  Acc  (\%)}               & \rotatebox{90}{98.00}                & \rotatebox{90}{78.13}                  & \rotatebox{90}{92.19}       & \rotatebox{90}{89.06}            & \rotatebox{90}{99.00}                & \rotatebox{90}{93.75}                 & \rotatebox{90}{71.88}            & \rotatebox{90}{98.44}       & \rotatebox{90}{97.00}                   & \rotatebox{90}{84.38}               & \rotatebox{90}{95.31}               & \rotatebox{90}{85.94}               & \rotatebox{90}{65.63}                  & \rotatebox{90}{89.06}                  & \rotatebox{90}{84.38}            \\ 
\hline
\end{tabular}
\end{center}

\caption{Accuracy per explainable feature of deep learning system using SAE}
\vspace{-1mm}
\end{table}

The two methods for inferring explanations from the features are DAAM and LAAM. The input to DAAM is the softmax layer of FLN whereas input to LAAM is the argmax of the softmax layer of FLN. DAAM uses the soft probabilities to compute the cosine similarity between the corresponding explainable features. During the evaluation phase the soft probabilities of each category for each sample are extracted and stored in memory, with the id of the writer and the name of the image as the keys. This data is then sorted in memory and $C_{sim}$ is computed. Once the scores are ready we  calculate $C_{OCS}$ for each sample. Next we compare the $C_{OCS}$ with $T$ to get the confusion metrics. With these confusion metrics we calculate the evaluation metrics Type 1 accuracy (T1), Type 2 accuracy (T2), Precision (P), Recall (R). Where, Intra Writer accuracy (Type 1) = $TP/S_w$, Inter Writer Accuracy (Type 2) = $FP/S_e$ and Overall Accuracy = $(TP+TN)/S_e$. Here, $S_e$ = Total number of samples in evaluation set, $S_w$ = Number of samples in resp. writer set. We repeat the same experiment for all the three datasets. Similar to DAAM, we have experimented LAAM using bayesian network on all the three datasets. The results are shown in Table 6. Output of the explainable model is shown in figure 4. Figure 4(a) shows an example of the explanations provided when the handwritten samples are written by different writer. The overall similarity score is low (0.3784) because the similarity between $is\_lowercase$, $staff\_of\_d$ and $exit\_stroke\_d$ is low. Similarly, Figure 4(b) shows an example of the explanations provided when the handwritten samples are written by the same writer.

\begin{table}[t]
\begin{center}
\begin{tabular}{|l|l|l|l|l|}
\hline
\textbf{Method}                        & \textbf{Metric}                      & \textbf{Seen}    & \textbf{Unseen}  & \textbf{Shuffled} \\ \hline
\multirow{3}{*}{DAAM\_CNN}          & Intra Writer Accuracy (Type 1)       & 86.34\%          & 66.12\%          & 76.77\%           \\ \cline{2-5} 
                                       & Inter Writer Accuracy (Type 2)       & 93.23\%          & 90.91\%          & 93.18\%           \\ \cline{2-5} 
                                       & Overall Accuracy                     & 92.64\%          & 91.51\%          & 94.87\%           \\ \hline
\multirow{3}{*}{DAAM\_AE}     & Intra Writer Accuracy (Type 1)       & 86.88\%          & 68.33\%          & 77.67\%           \\ \cline{2-5} 
                                       & Inter Writer Accuracy (Type 2)       & 94.80\%          & 92.41\%          & 94.13\%           \\ \cline{2-5} 
                                       & Overall Accuracy                     & 94.58\%          & 92.31\%          & 94.34\%           \\ \hline
\multirow{3}{*}{DAAM\_SAE} & Intra Writer Accuracy (Type 1)       & 88.12\%          & 70.82\%          & 80.98\%           \\ \cline{2-5} 
                                       & Inter Writer Accuracy (Type 2)       & 95.58\%          & 93.49\%          & 95.08\%           \\ \cline{2-5} 
                                       & Overall Accuracy                     & \textbf{95.78\%} & \textbf{94.27\%} & \textbf{95.23\%}  \\ \hline
\multirow{3}{*}{LAAM\_SAE}                  & Intra Inter Writer Accuracy (Type 1) & 75.17\%          & 71.92\%          & 81.67\%             \\ \cline{2-5} 
                                       & Inter Writer Accuracy (Type 2)       & 95.77\%          & 87.30\%          & 93.90\%           \\ \cline{2-5} 
                                       & Overall Accuracy                     & 85.15\%          & 85.79\%          & 92.76\%             \\ \hline
\end{tabular}
\end{center}
\caption{Performance using DAAM and LAAM on XAI-AND dataset. Feature learning using translation-invariant skip-auto-encoder outperforms basic CNN and basic auto-encoder}
\vspace{-2mm}
\end{table}

\section{Conclusion}
We have made XAI-AND dataset publicly available. The dataset contains explainable features (meta-data) associated with each of the handwritten sample in the dataset. Furthermore, we provide an effective deep learning system which is capable of learning and generating explanations for the task of handwriting verification. Our experiments denote that threshold based methods like cosine similarity and likelihood ratio can be used to explain the output of deep learning system. We show that deep learning model can be amalgamated with Bayesian models to provide useful explanations for FDE. Currently our system is two pass, in the first pass we train the deep learning model and in the second pass we train the Bayesian network. However, we plan to make the system end to end trainable in the future. Moreover, we plan to build a pipeline to update model parameters while the FDEs are labeling the dataset and provide suggestive feedbacks on their input. Finally, our future plan is to provide visual explanations using GradCAM \cite{selvaraju2017grad} like approaches. 

\bibliography{egbib}
\end{document}